\let\accentvec\vec
\documentclass[runningheads,a4paper]{llncs}

\let\vec\accentvec

\usepackage{amssymb}
\usepackage{graphicx}

\usepackage{amsmath}

\usepackage{url}
\urldef{\mailsa}\path|{alfred.hofmann, ursula.barth, ingrid.haas, frank.holzwarth,|
\urldef{\mailsb}\path|anna.kramer, leonie.kunz, christine.reiss, nicole.sator,|
\urldef{\mailsc}\path|erika.siebert-cole, peter.strasser, lncs}@springer.com|
\newcommand{\keywords}[1]{\par\addvspace\baselineskip
\noindent\keywordname\enspace\ignorespaces#1}

\DeclareMathOperator{\ocap}{\displaystyle{\small{\textcircled{{\scriptsize $\cap$}}}}}
\graphicspath{{./fig1/}}
\usepackage{amsmath}
\begin{document}

\mainmatter  

\title{Evidential-EM algorithm applied to progressively censored observations}

\titlerunning{Evidential-EM algorithm applied to progressively censored observations}

%
%
\author{Kuang Zhou\inst{1,2} \and Arnaud Martin\inst{2}
 \and Quan Pan\inst{1}}
\authorrunning{Kuang Zhou et al.} 
%
%
\institute{School of Automation, Northwestern Polytechnical University, \\ Xi'an, Shaanxi 710072, PR China
\and
IRISA, University of Rennes 1, Rue E. Branly, 22300 Lannion, France
\\
kzhoumath@163.com,  Arnaud.Martin@univ-rennes1.fr, quanpan@nwpu.edu.cn}

\toctitle{Lecture Notes in Computer Science}
\tocauthor{Authors' Instructions}
\maketitle

\begin{abstract}
Evidential-EM (E2M) algorithm is an effective approach for computing maximum likelihood estimations under finite  mixture models, especially when there is uncertain information about data. In this paper we present an extension of the E2M method in a particular case of incomplete data, where the loss of information is due to both  mixture models and censored observations. The prior uncertain information is expressed by belief functions, while the  pseudo-likelihood function is derived based on  imprecise observations and prior knowledge.  Then E2M method is evoked to maximize the generalized likelihood function to obtain the optimal estimation of parameters. Numerical examples show that the proposed method could effectively integrate the uncertain prior information with the current imprecise knowledge conveyed by the observed data.

\end{abstract}

\section{Introduction}

In life-testing experiments, the data are often censored. A datum $T_i$
is said to be right-censored if the event occurs at a time after
a right bound, but we do not exactly know when. The only information we have is
this right bound. 
Two most common right censoring schemes are termed as
Type-I and Type-II censoring. The experiments using these test schemes
have the drawback that they do not allow
removal of samples at time points other than the terminal of the experiment. The progressively censoring scheme, which possesses this advantage, has become very popular in the life tests in the last few years~\cite{pc}. The censored data provide some kind of imprecise information for reliability analysis.

It is interesting to evaluate the reliability performance for items with mixture distributions. When the
population is composed of several subpopulations, an instance in the data set is expected to have
a label which represents the origin,
that is, the subpopulation from which the data is observed. In real-world data, observed labels may carry only partial information about the origins of samples. Thus there are concurrent imprecision and uncertainty for the censored data from mixture distributions.
The Evidential-EM (E2M) method, proposed by Den{\oe}ux~\cite{e2m,e2m1}, is an effective approach for computing maximum likelihood estimates for the mixture problem, especially when there is both imprecise and uncertain knowledge about the data. However, it has not been used for reliability analysis and the censored life tests.

This paper considers a special kind of incomplete data in life tests, where the loss of information is due simultaneously to the  mixture problem  and to censored observations. The data set analysed in this paper is merged by samples from different classes. Some uncertain information about class values of these unlabeled data is expressed by belief functions.
The pseudo-likelihood function is obtained based on the imprecise observations and uncertain prior information, and then E2M method is invoked to maximize the generalized likelihood function. The simulation studies show that  the proposed method could take advantages of using
the partial labels, and thus incorporates more information than traditional EM algorithms.



\section{Theoretical analysis}
Progressively censoring scheme has attracted considerable attention in recent years, since it has the flexibility of allowing removal of units at points other than the terminal point of the experiment~\cite{pc}. The theory of belief functions is first described by Dempster~\cite{ds1} with the study of upper and lower probabilities and extended by Shafer later~\cite{ds2}. This section will give a brief description of these two concepts.
\subsection{The Type-II progressively censoring scheme}
\label{PCS}
The model of Type-II progressively censoring scheme (PCS) is described as follows~\cite{pc}. Suppose $n$ independent identical items are placed on a life-test with the corresponding lifetimes $X_1,X_2,\cdots,X_n$
being identically distributed. We assume that $X_i$ $(i=1,2,\cdots,n)$ are i.i.d. with probability density function (pdf) $f(x;\theta)$ and cumulative distribution function (cdf) $F(x;\theta)$. The integer $J<n$ is fixed at the beginning
of the experiment. The values $R_1,R_2,\cdots,R_J$ are $J$ pre-fixed satisfying $R_1+R_2+\cdots+R_J+J=n$.
During the experiment, the $j^{th}$ failure is observed and immediately after the failure, $R_j$
functioning items are
randomly removed from the test. We denote the time of the $j^{th}$ failure by $X_{j:J:n}$, where $J$ and $n$ describe the censored scheme used in the experiment, that is, there are $n$ test units and the experiment stops after $J$ failures are observed.
Therefore, in the presence of Type-II progressively censoring schemes, we have  the observations $\{X_{1:J:n},\cdots,X_{J:J:n}\}$. The likelihood function can be given by
\begin{equation}
\label{lilelihood}
L(\theta;x_{1:J:n},\cdots,x_{J:J:n}) =C \prod_{i=1}^J f(x_{i:J:n};\theta)[1-F(x_{i:J:n};\theta)]^{R_i}, 
\end{equation}
where $C=n(n-1-R_1)(n-2-R_1-R_2)\cdots(n-J+1-R_1-R_2-\cdots-R_{J-1})$.

\subsection{Theory of belief functions}
Let $\Theta=\{\theta_1,\theta_2, \ldots, \theta_N\}$ be the finite domain of $X$, called the discernment frame. The mass function is defined on the power set $2^\Theta=\{A:A\subseteq \Theta\}$.
The function $m:2^\Theta\rightarrow [0,1]$ is said to be the basic belief assignment (bba) on $\text{2}^{\Theta} $, if it satisfies:
 \begin{equation}
 \sum_{A\subseteq \Theta}m(A)=1.
 \end{equation}
Every $A\in 2^\Theta$ such that $m(A)>0$ is called a focal element. The credibility and plausibility functions are  defined in Eq.~\eqref{bel} and Eq.~\eqref{pl}.
  \begin{equation}
  Bel\text{(}A\text{)}=\sum_{\emptyset \neq B\subseteq A} m\text{(}B\text{)}, \forall A\subseteq \Theta,
  \label{bel}
  \end{equation}
   \begin{equation}
   Pl\text{(}A\text{)}=\sum_{B\cap A \neq \emptyset}  m\text{(}B\text{)}, \forall A\subseteq \Theta .
   \label{pl}
   \end{equation}
Each quantity $Bel(A)$ denotes the degree to which the evidence supports $A$, while $Pl(A)$ can be interpreted as an upper bound on the degree of support that could be assigned to $A$ if more specific information became available~\cite{TBM}. The function $pl:\Theta \rightarrow [0,1]$ such that $pl(\theta) = Pl(\{\theta\})$ is called the contour function associated to $m$.

If $m$ has a single focal element $A$, it is said to be categorical and denoted as $m_A$. If all focal elements of $m$ are singletons, then $m$ is said to be Bayesian. Bayesian mass functions are  equivalent to probability distributions.

If there are two distinct pieces of evidences (bba) on the same frame, they can be combined using Dempster's rule~\cite{ds2} to form a new bba:
\begin{equation}
m_{1 \oplus 2}(C)=
\frac{\sum_{A_i \cap  B_j=C}m_1(A_i)m_2(B_j)}{\text{1}-k} \qquad  \forall C \subseteq \Theta,C \neq \emptyset \\ 
\end{equation}

If $m_1$ is Bayesian mass function,and its corresponding contour function is  $p_1$. Let $m_2$ be an arbitrary mass function with contour function $pl_2$. The combination of $m_1$ and $m_2$ yields a Bayesian mass function
$m_1 \oplus m_2$
with contour
function $p_1\oplus pl_2$
defined by
\begin{equation}
p_1\oplus pl_2=\frac{p_1(\omega)pl_2(\omega)}{\sum _{\omega ^{'} \in \Omega} p_1(\omega^{'})pl_2(\omega^{'})}.
\end{equation}
The conflict between $p_1$ and $pl_2$ is $k=1-\sum _{\omega^{'} \in \Omega} p_1(\omega^{'})pl_2(\omega^{'})$.
It equals one minus the expectation of $pl_2$ with respect to $p_1$.

\section{The E2M algorithm for Type-II PCS}

\subsection{The generalized likelihood function and E2M algorithm}
E2M algorithm, similar to the EM method, is an iterative optimization tactics to obtain the maximum of the observed likelihood function~\cite{e2m,e2m1}. However, the data applied to E2M model can be imprecise and uncertain. The imprecision may be brought by missing information or hidden variables, and this problem
can be solved by the EM approach. The uncertainty may be due to the unreliable sensors, the errors caused by the measuring or estimation methods and so on. In the E2M model, the uncertainty is represented by belief functions.

Let $X$ be a discrete variable defined on $\Omega_X$ and  the probability density function is $p_X(\cdot;\theta)$. If $x$ is an observation sample of $X$, the likelihood function can be expressed as:
\begin{equation}
L(\theta;x)=p_X(x;\theta).
\end{equation}
If $x$ is not completely observed, and what we only know is that $x \in A,A \subseteq \Omega_X$, then the likelihood function becomes:
\begin{equation}
L(\theta;A)=\sum_{x \in A}p_X(x;\theta).
\end{equation}
If there is some uncertain information about $x$, for example, the experts may give their belief about $x$ in the form of mass functions: $m(A_i),i=1,2,\cdots,r,$ $A_i \subseteq \Omega_X$, then the likelihood becomes:
\begin{equation}
L(\theta;m)=\sum_{i=1}^r m(A_i)L(\theta;A_i)=\sum_{x \in \Omega_x}p_X(x;\theta)pl(x).
\label{general likelihood}
\end{equation}
It can be seen from Eq.~\eqref{general likelihood} that the likelihood $L(\theta;m)$ only depends on $m$ through its associated contour function $pl$. Thus we could write indifferently $L(\theta;m)$ or $L(\theta;pl)$.

Let $W=(X,Z)$ be the complete variable set. Set $X$ is the observable data while $Z$ is unobservable but with some uncertain knowledge in the form of $pl_Z$. The log-likelihood based on the complete sample is $\log L(\theta;W)$. In E2M,  the observe-data log likelihood is $\log L(\theta;X,pl_Z)$.

In the E-step of the E2M algorithm, the pseudo-likelihood function should be calculated as:
\begin{equation}
Q(\theta, \theta^k)=\mathrm{E}_{\theta ^k}[\log L(\theta;W)|X,pl_Z; \theta^k],
\end{equation}
where $pl_Z$ is the contour function describing our uncertainty on $Z$, and $\theta^k$ is the parameter vector obtained at the $k^{th}$ step. $\mathrm{E}_{\theta ^k}$ represents the expectation with respect to the following density:
\begin{equation}
\gamma ^{'}(Z=j|X,pl_Z;\theta^k)\triangleq  \gamma(Z=j|X;\theta^k)\oplus pl_Z.
\end{equation}
Function $\gamma ^{'}$ could be regarded as a combination of conditional probability density \linebreak $\gamma(Z=j|X;\theta^k)=p_Z(Z=j|X;\theta^k)$ and the contour function $pl_Z$.
It depicts the current information based on the observation $X$ and  the prior uncertain information on $Z$, thus this combination is similar to the Bayes rule.

According to the Dempster combination rule and Eq.~\eqref{general likelihood}, we can get:
\begin{equation}
\gamma ^{'} (Z=j|X,pl_Z;\theta^k)=\frac{r(Z=j|X;\theta^k)pl_Z(Z=j)}{\sum_j r(Z=j|X;\theta^k)pl_Z(Z=j)}.
\end{equation}
Therefore, the pseudo-likelihood is:
\begin{equation}
Q(\theta, \theta^k)=\frac{\sum_j r(Z=j|X;\theta^k)pl(Z=j) \log L(\theta;W)}{L(\theta^k;X,pl_Z)}.
\end{equation}

The M-step is the same as EM and requires the maximization of $Q(\theta, \theta^k)$ with respect to $\theta$. The E2M algorithm alternately repeats the E- and M-steps above until the increase of general observed-data likelihood becomes smaller than a given threshold.

\subsection{Mixed-distributed progressively censored data }
Here, we present a special type of incomplete data, where the imperfection of
information is due both to the  mixed-distribution  and to some censored observations. Let $Y$ denote the lifetime of  test samples. The $n$ test samples can de divided into two parts, {\em i.e.} $Y_1,Y_2$, where $Y_1$ is the set of observed data, while $Y_2$ is the censored data set. Let $Z$ be the class labels and $W=(Y,Z)$ represent the complete data.

Assume that $Y$ is from mixed-distribution with p.d.f.
\begin{equation}
f_Y(y;\theta)=\sum_{z=1}^{p} \lambda_z f(y;\xi_z),
\label{mix_dis}
\end{equation}
where $\theta=(\lambda_1,\cdots,\lambda_p,
\xi_1,\cdots,\xi_p)$.
The complete data distribution of $W$ is given by $P(Z=z)=\lambda_z$ and $P(Y|Z=z)=f(y;\xi_z)$. Variable
$Z$ is hidden but we can have a prior knowledge about it. 
This kind of prior uncertain information of $Z$ can be described in the form of belief functions:
\begin{equation}
pl_Z(Z=j)=pl_j,j=1,2,\cdots,p.
\end{equation}

The likelihood of the complete data is:
\begin{equation}
L^{c}(\theta;Y,Z)=\prod_{j=1}^{n}f(y_{j},z_{j};\theta),
\end{equation}
and the pseudo-likelihood function is:
\begin{equation}
Q(\theta,\theta^{k})=\mathrm{E}_{\theta^k}[\log L{}^{c}\text{(\ensuremath{\theta;Y,Z}})|Y^{*},pl_Z;\theta^{k}],
\end{equation}
where $\text{\ensuremath{\mathrm{E}}}_{\theta^k}[\cdot|Y^{*},pl_Z;\theta^k]$
denotes expectation with respect to the conditional distribution of $W$ given the observation $Y^{*}$ and the uncertain information $pl_Z$.
\begin{theorem}
For $(y_{j}$,$z_{j})$ are complete and censored,
$\text{\ensuremath{f_{YZ}(y_{j},z_{j}|y_{j}^{*};\theta^{k})}}$ can be calculated according to Eq.~\eqref{eq:fyz1} and Eq.~\eqref{eq:fyz2} respectively.  Let $y_{j}^*$ be the $j^{th}$ observation. If the $j^{th}$ sample is completely observed, $y_{j}=y_{j}^*$; Otherwise $y_{j} \geq y_{j}^*.$
\begin{equation}
\text{\ensuremath{f_{YZ}^{1}(y_{j},z_{j}|y_{j}^{*};\theta^{k})=\mathrm{I}_{\{y_{j}=y_{j}^{*}\}}P_{1jz}^{k}},}
\label{eq:fyz1}
\end{equation}
\begin{equation}
\text{\ensuremath{f_{YZ}^{2}(y_{j},z_{j}|y_j^{*};\theta^{k})=
\mathrm{I}_{\{y_{j}>y_{j}^{*}\}}P_{2jz}^{k}\frac{f(y_{j};\xi_{z}^{k})}
{\overline{F}(y_{j}^{*};\xi_{z}^{k})}}.}\label{eq:fyz2}
\end{equation}
where $P_{1jz}^k$ and $P_{2jz}^k$ are shown in Eq.~\eqref{pjzk}.
\begin{equation}
P_{jz}^{k}(z_{j}=z|Y^{*};\theta)=\begin{cases}
P_{1jz}^{k}(z_{j}=z|y_{j}^{*};\theta^{k}) & \text{for the completely observed data}\\
P_{2jz}^{k}(z_{j}=z|y_{j}^{*};\theta^{k}) & \text{for the censored data}
\end{cases}
\label{pjzk}
\end{equation}
where
\begin{equation}
P_{1jz}^{k}(z_{j}=z|y_{j}^{*};\theta^{k})=
\frac{f(y_{j}^{*};\xi_{z}^{k})\lambda_{z}^{k}}{\sum_{z}f(y_{j}^{*};\xi_{z}^{k})\lambda_{z}^{k}},
\label{p1jz}
\end{equation}
\begin{equation}
P_{2jz}^{k}(z_{j}=z|y_{j}^{*};\theta^{k})=
\frac{\overline{F}(y_{j}^{*};\xi_{z}^{k})
\lambda_{z}^{k}}{\sum_{z}\text{\ensuremath{\overline{F}}}(y_{j}^{*};\xi_{z}^{k})\lambda_{z}^{k}}.
\label{p2jz}
\end{equation}
\begin{proof}
If ($y_{j}$,$z_{j}$) are completely observed,
$$\text{\ensuremath{f_{yz}^{1}(y_{j},z_{j}|y_{j}^{*};\theta^{k})
=P_{1jz}^{k}f(y_{j}|y_{j}^{*}=y_{j},Z_{j}=z;\theta^{k})}},$$
we obtain Eq.~\eqref{eq:fyz1}.

If ($y_{j}$,$z_{j}$) are censored,
$$\text{\ensuremath{f_{yz}^{2}(y_{j},z_{j}|y_{j}^{*};\theta^{k})
=P_{2jz}^{k}f(y_{j}|y_{j}^{*} < y_{j},Z_{j}=z;\theta^{k})}},$$

From the theorem in \cite{EM1},
$$\text{\ensuremath{f(y_{j}|y_{j}^{*} < y_{j},Z_{j}=z;\theta^{k})}}
=\frac{f(y_{j};\xi_{z}^{k})}{\overline{F}(y_{j}^{*};\xi_{z}^{k})}\text{I}_{\{y_{j}>y_{j}^{*}\}},$$
we can get Eq.~\eqref{eq:fyz2}.

This completes this proof.
\end{proof}
\end{theorem}

From the above theorem, the pseudo-likelihood function can be written as:

\begin{equation}
\begin{aligned}
Q(\theta, \theta^{k})&= \mathrm{E}_{\theta^k}[\log f{}^{c}\text{(\ensuremath{Y,Z}})|Y^{*},pl_Z;\theta^{k}]\\
&= \sum_{j=1}^{n}  \mathrm{E}_{\theta^k}[\log\lambda_{z}+\log f(y_{j}|\xi_{z})|Y^{*},pl_Z;\theta^{k}]
\\
&=\sum_{y_j \in Y_1}\sum_{z}P_{1jz}^{'k}\log\lambda_{z}+\sum_{y_j \in Y_2}\sum_{z}P_{2jz}^{'k}\log\lambda_{z}\\
&+\sum_{y_j \in Y_1}\sum_{z}P_{1jz}^{'k}\log f(y_{j}^{*}|\xi_{z})\\
&+\sum_{y_j \in Y_2} \sum_{z}P_{2jz}^{'k}\int_{y_{j}^{*}}^{+\infty}\log f(x|\xi_{z})\frac{f(x|\ensuremath{\xi_{z}^{k}})}{\overline{F}(y_{j}^{*};\xi_{z}^{k})}\mathrm{d}x,
\label{gene_per_likeli}
\end{aligned} \end{equation}
where
$$
P_{ijz}^{'k}(z_{j}=z|y_j^{*},pl_{Z_j};\theta^k)=P_{ijz}^{k}(z_{j}=z|y_{j}^{*};\theta^{k})\oplus pl_{Z_j}, i=1,2.
\label{pjzk2}
$$
It can be seen that $P_{ijz}^{'k}(z_{j}=z|y_j^{*},pl_{Z_j};\theta^k)$ is a Dempster combination of the prior  and the observed information.

Assume that the data is from the mixed-Rayleigh distribution without loss of generality, the p.d.f. is shown in  Eq.~\eqref{mexp}:
\begin{equation}
f_X(x;\lambda,\xi)=\sum_{j=1}^{p}\lambda_{j} g_X(x;\xi_j)=\sum_{j=1}^{p}\lambda_{j}\xi_{j}^2 \exp\{-\frac{1}{2}\xi_{j}^2 x^2\},
\label{mexp}
\end{equation}

After the $k^{th}$ iteration  and
$\text{\ensuremath{\theta}}^{k}=\lambda^{k}$ is got, the $(k+1)^{th}$ step of E2M algorithm is shown as follows:
\begin{enumerate}
\item \textbf{E-step:} For $j=1,2,\cdots,n$, $z=1,2\cdots,p$, use Eq.~\eqref{gene_per_likeli} to obtain the conditional p.d.f. of $\log L^c(\theta;W)$ based on the observed data, the prior uncertain information and the current parameters.
\item \textbf{M-step:} Maximize $Q(\theta|\theta^{k})$ and update the parameters:
\begin{equation}
\text{\ensuremath{\lambda}}_{z}^{k+1}=\frac{1}{n}\left(\sum_{y_j\in Y_1}P_{1jz}^{'k}+\sum_{y_j\in Y_2}P_{2jz}^{'k}\right),
\end{equation}

\begin{equation}
(\xi_{z}^{k+1})^2=\frac{2\left( \sum_{y_j \in Y_1} P_{1jz}^{'k}+\sum_{y_j \in Y_2} P_{2jz}^{'k}\right)}{\sum_{y_j \in Y_1}P_{1jz}^{'k}y_{j}^{*^2}
+\sum_{y_j \in Y_2}P_{2jz}^{'k}(y_{j}^{*^2}+2/(\xi_{z}^{k})^2)}.
\end{equation}
\end{enumerate}

It should be pointed out that the maximize of  $Q(\theta, \theta^{k})$ is conditioned on $\sum_{i=1}^p \lambda_i=1.$ By Lagrange multipliers method we have the new objective function: $$Q(\theta,\theta^{k})-\alpha (\sum_{i=1}^p \lambda_i-1).$$

\section{Numerical results}
In this section, we will use Monte-Carlo method to test the proposed method. The simulated data set in this section is  drawn from mixed Rayleigh distribution as shown in Eq.~\eqref{mexp} with $p=3$, $\lambda=(1/3,1/3,1/3)$ and  $\xi=(4,0.5,0.8)$.
The test scheme is $n=500$, $m=n*0.6$, $R=(0,0,\cdots,n-m)_{1 \times m}$. Let the initial values  be $\lambda^0=(1/3,1/3,1/3)$ and $\xi^0=(4,0.5,0.8)-0.01$.
As mentioned before, usually there is no information about the subclass labels of the data, which is the case of unsupervised learning. But in real life, we may get some prior uncertain knowledge from the experts or experience. These partial information is assumed to be in the form of belief functions here.

\begin{center}
\begin{figure}[!thbt]
\centering
	\includegraphics[width=0.45\linewidth]{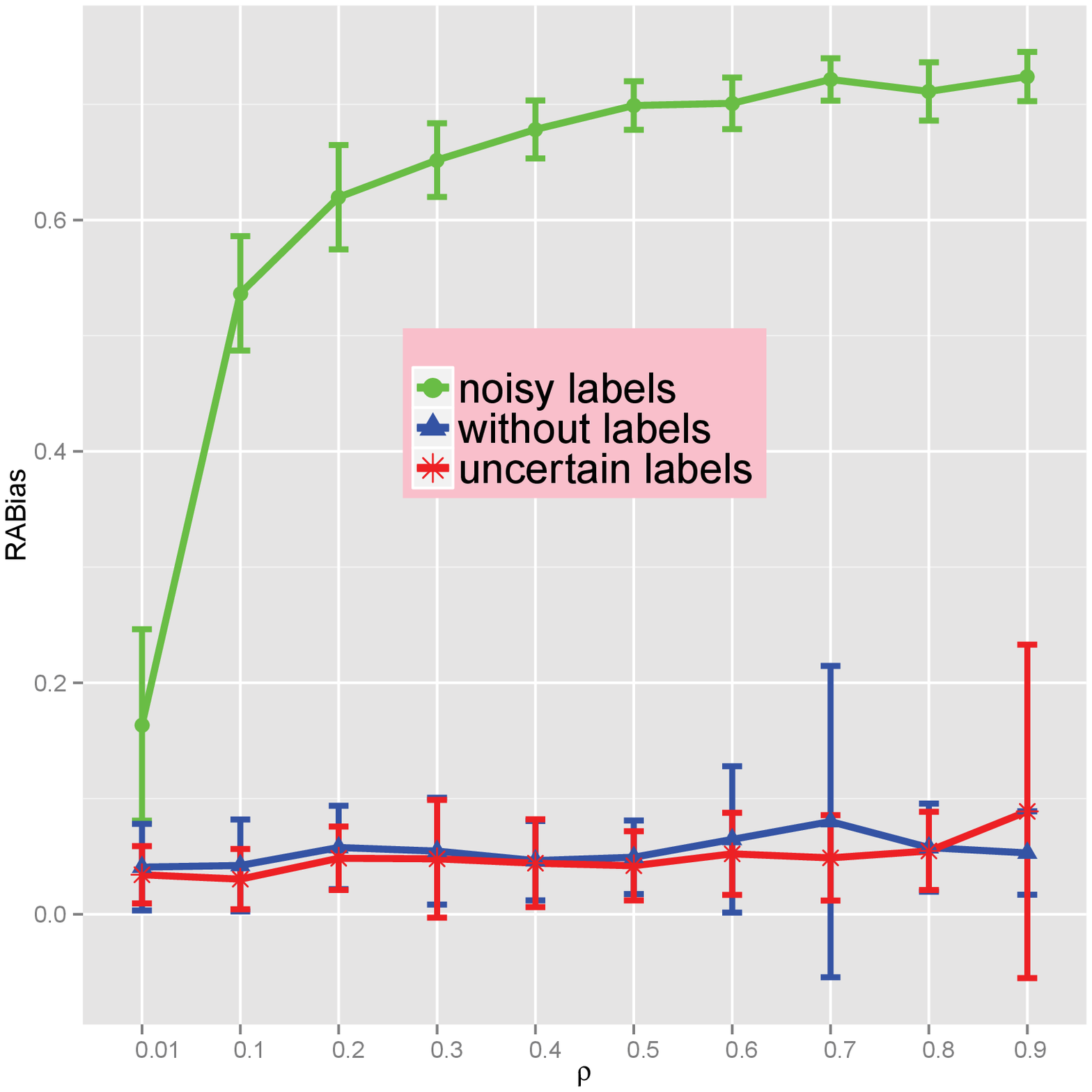}
	\hfill
	\includegraphics[width=.45\linewidth]{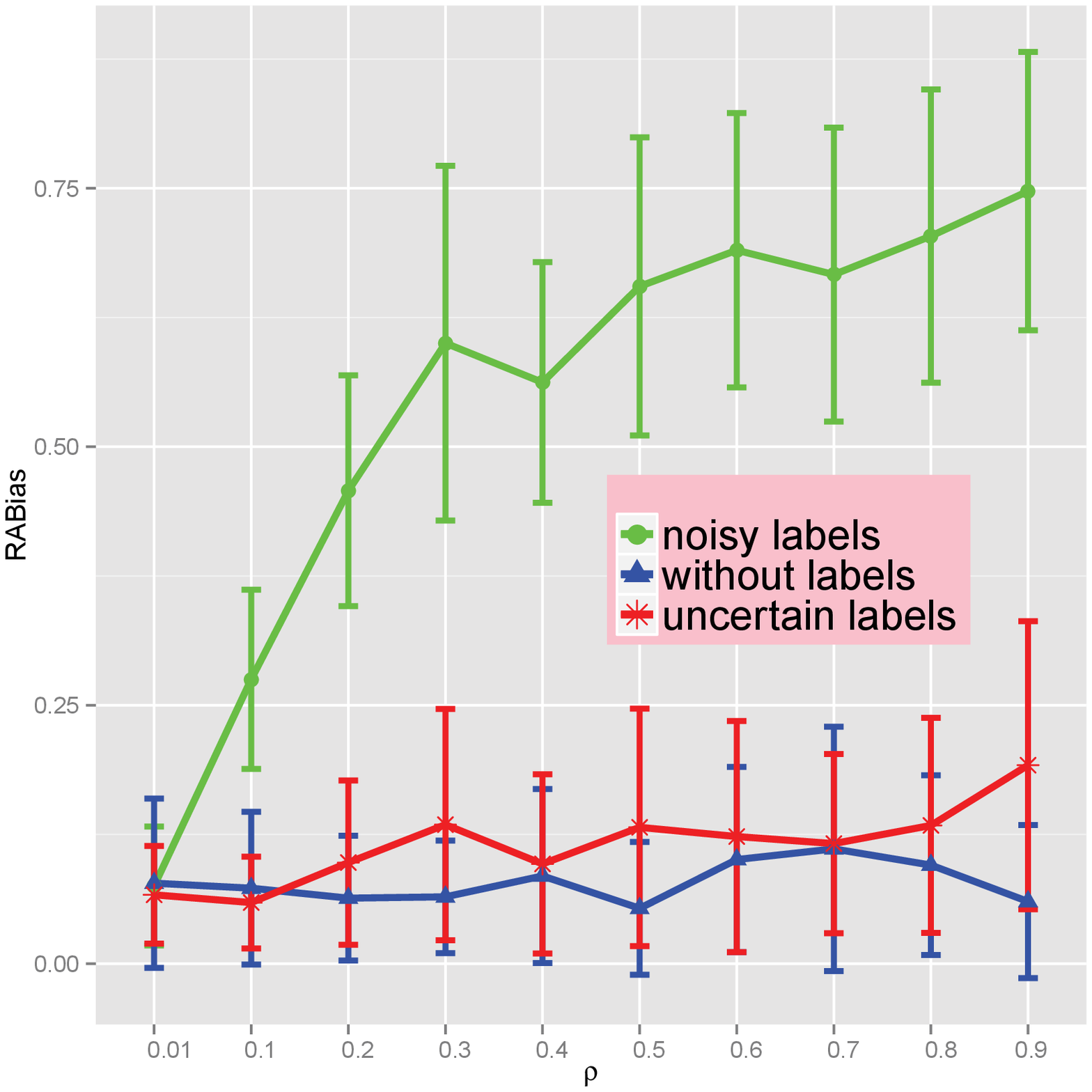}
	\hfill
 \parbox{.45\linewidth}{\centering\small a. Estimation of $\xi_1$}
	\hfill
	\parbox{.45\linewidth}{\centering\small b. Estimation of $\xi_2$}
	\hfill
 \includegraphics[width=.45\linewidth]{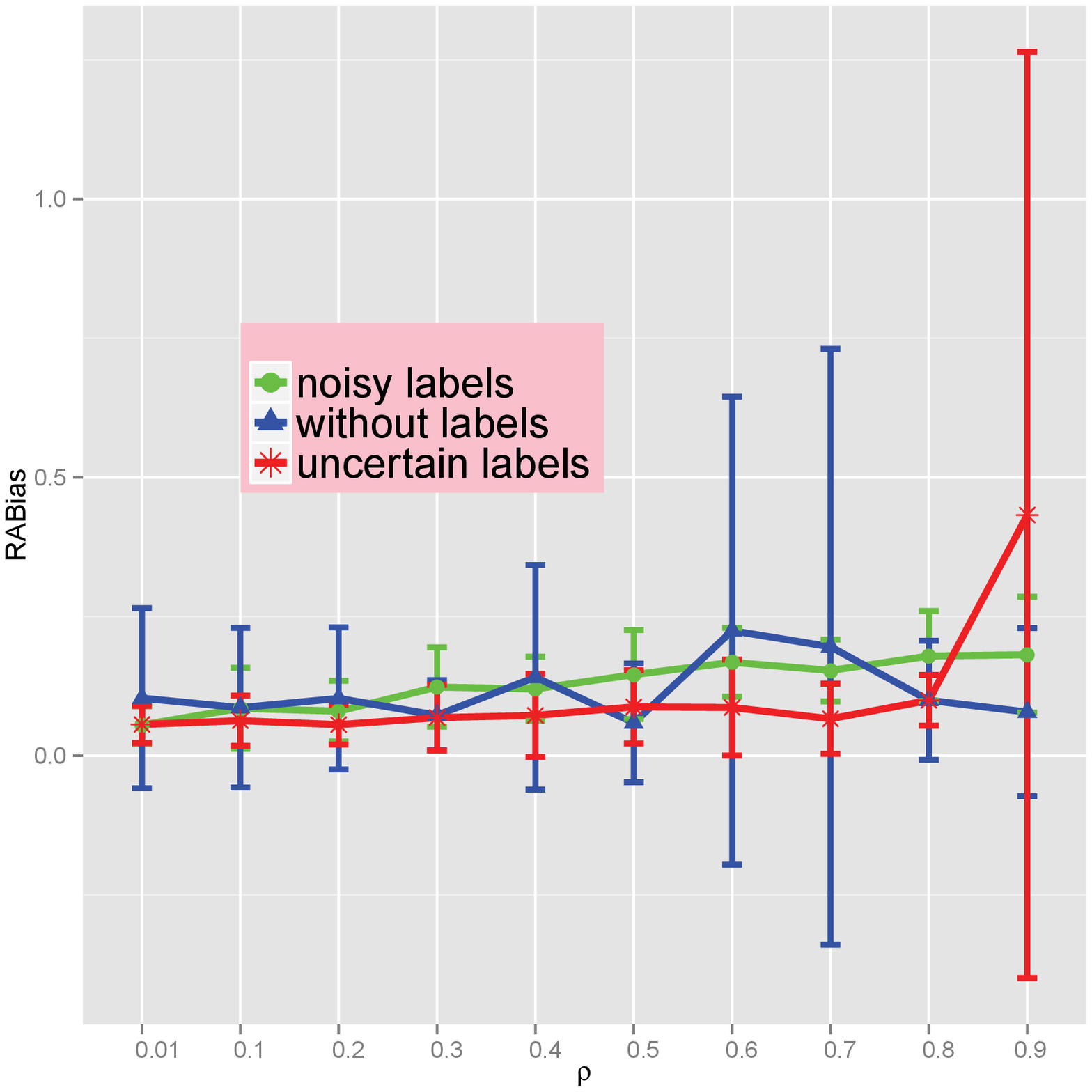}
     \hfill
    \parbox{.8\linewidth}{\centering\small c. Estimation of $\xi_3$}
	\hfill
	\caption{Average RABias values (plus and minus one standard deviation) for 20 repeated experiments, as a function of the error probability $\rho$ for the simulated labels.}
\label{newfig}
\end{figure}
\end{center}

\vspace{-3em}
To simulate the uncertainty on the labels of the data, the original generated datasets are corrupted as follows. For each data $j$, an error probability $q_j$ is drawn randomly from a beta distribution with mean $\rho$ and standard deviation 0.2. The value $q_j$ expresses the doubt by experts on the class of sample $j$. With probability $q_j$, the label of sample $j$ is changed to any (three) class (denoted by $z_j^*$) with equal probabilities. The plausibilities are then determined as
\begin{equation}
  pl_{Z_j}(z_j)=\begin{cases}
    \frac{q_j}{3} & ~~~if ~~z_j\neq z_j^*,\\
    \frac{q_j}{3}+1-q_j &~~~ if ~~ z_j=z_j^*
  \end{cases}.
\end{equation}
The results of our approach with uncertain labels are compared with the cases of noisy labels and no information on labels. The former case with noisy labels is like supervised learning, while the latter is the traditional EM algorithm applied to progressively censored data. In each case, the E2M (or EM) algorithm is run 20 times. The estimations of parameters are compared to their real value using absolute relative bias (RABias).
We recall that this commonly used measure equals 0 for the absolutely exact estimation $\hat{\theta}=\theta$.
\begin{center}
\begin{figure}[!thbt]
\centering
	\includegraphics[width=0.45\linewidth]{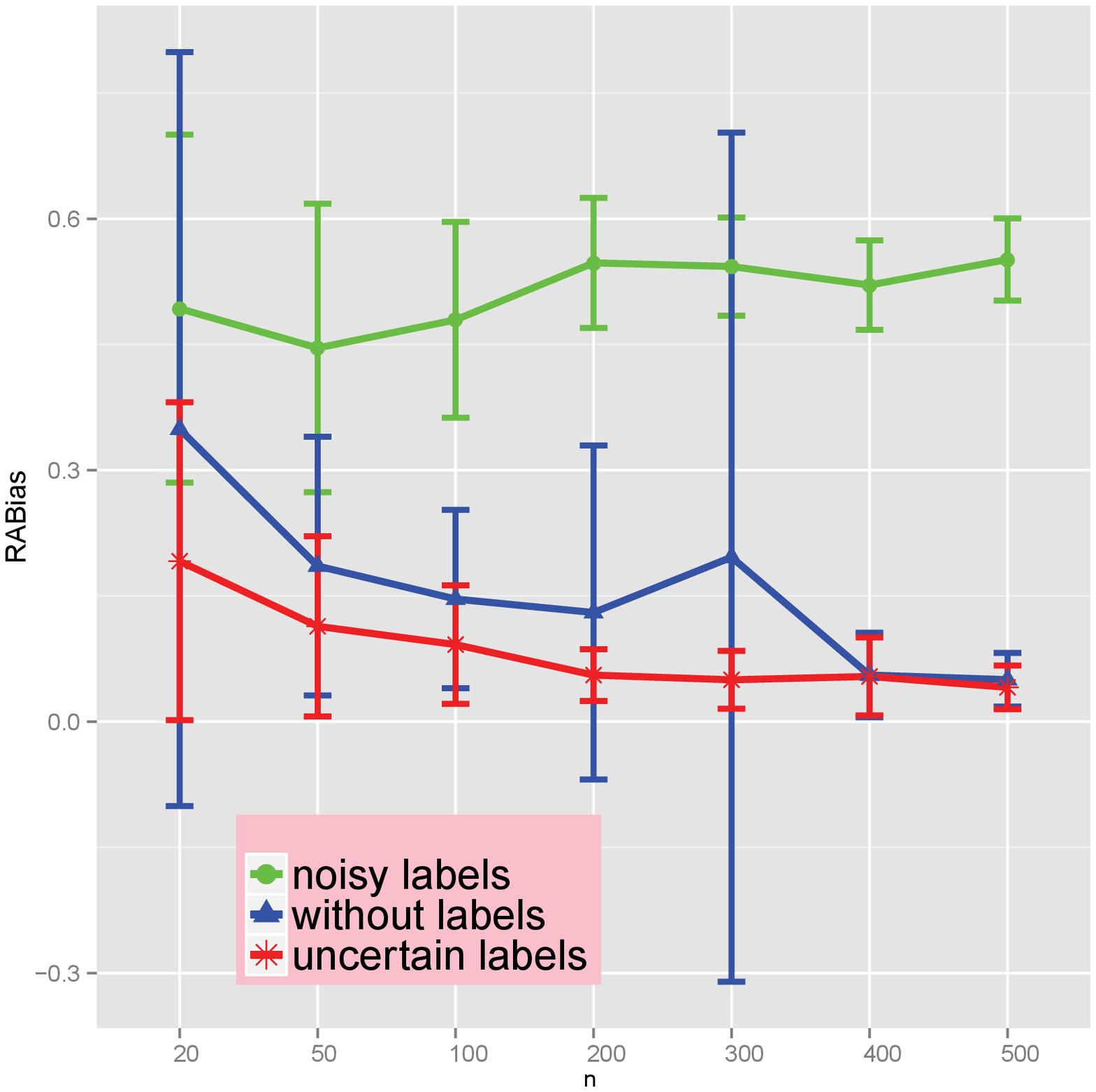}
	\hfill
	\includegraphics[width=.45\linewidth]{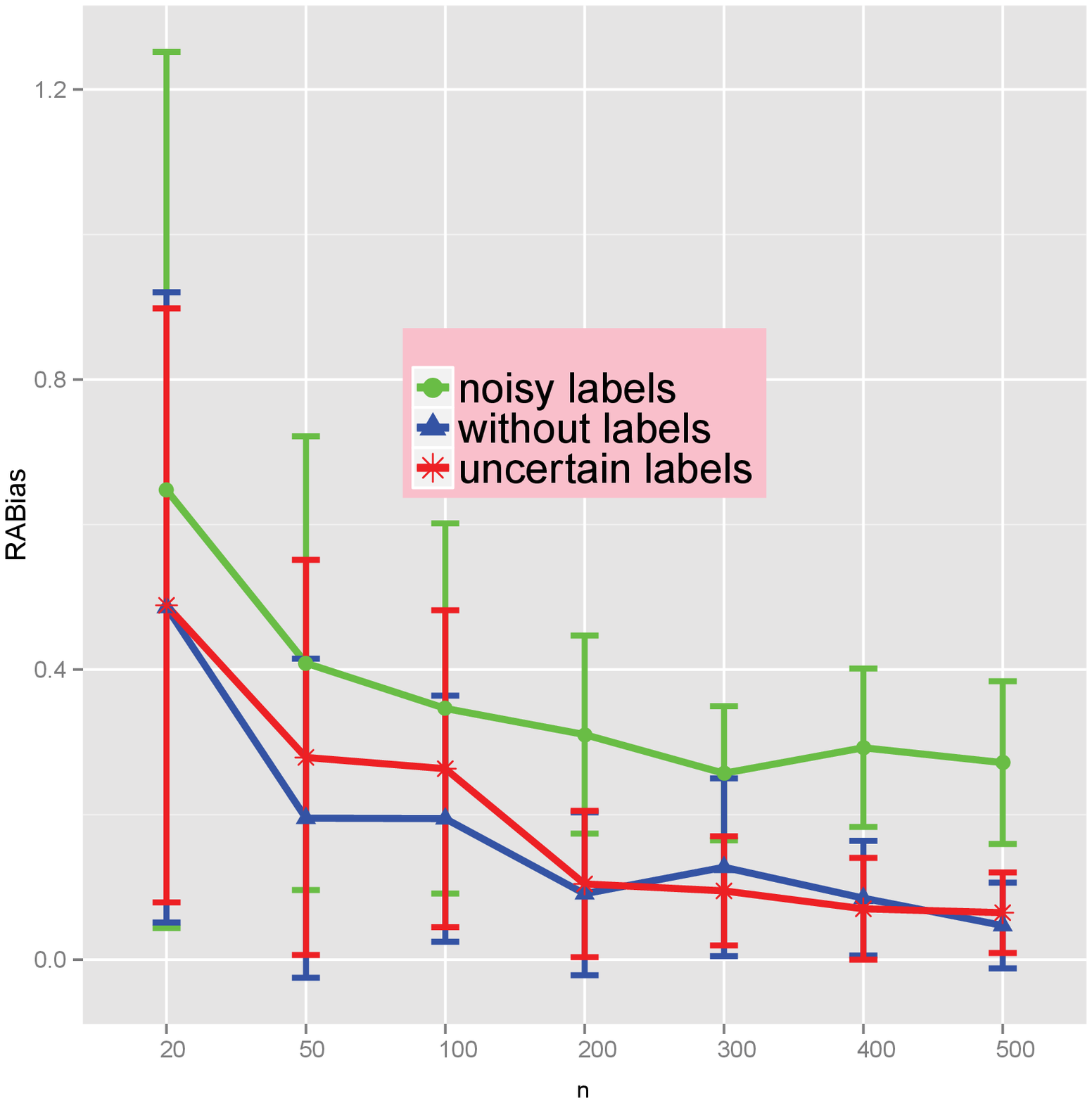}
	\hfill
 \parbox{.45\linewidth}{\centering\small a. Estimation of $\xi_1$}
	\hfill
	\parbox{.45\linewidth}{\centering\small b. Estimation of $\xi_2$}
	\hfill
 \includegraphics[width=.45\linewidth]{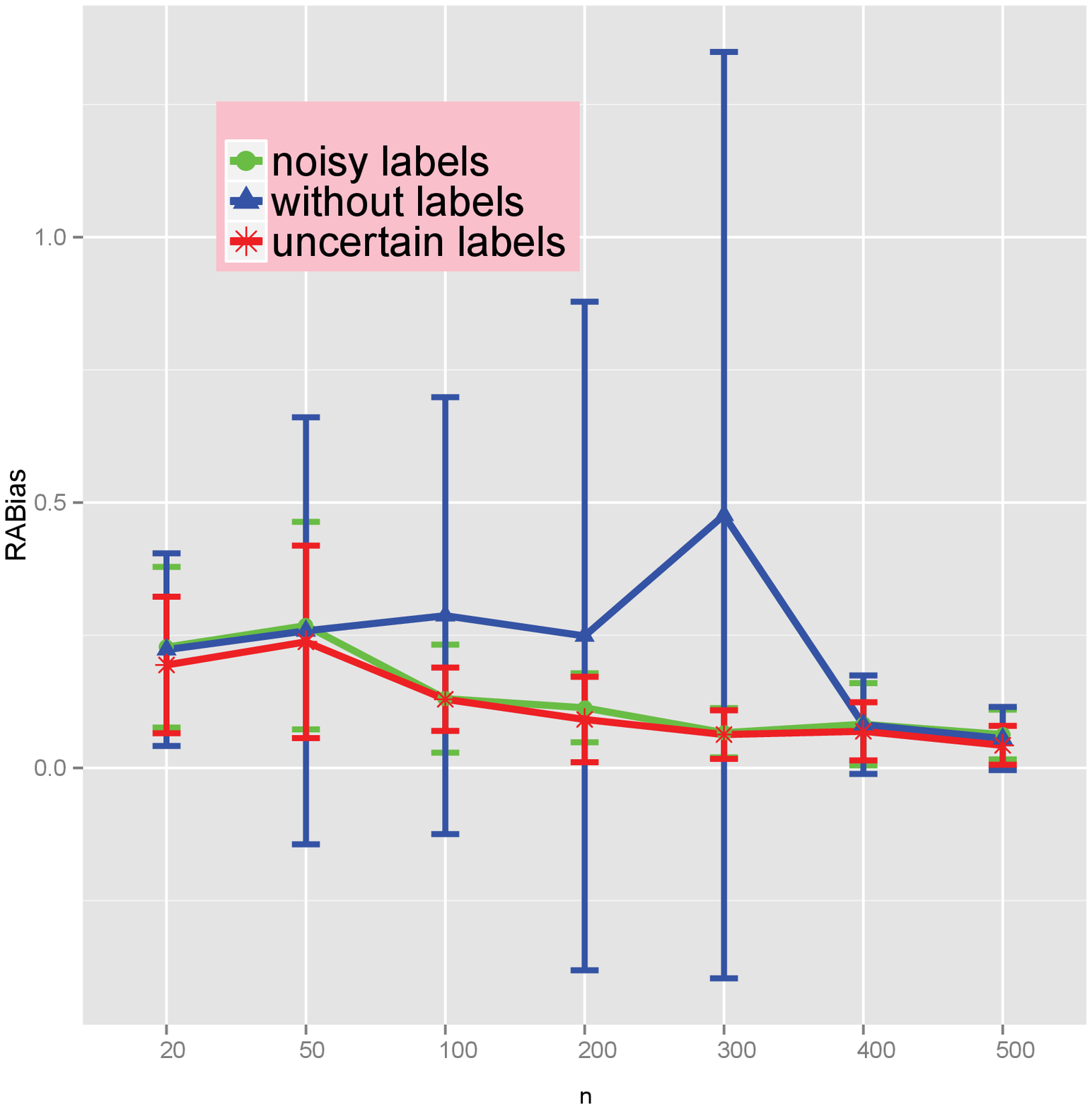}
     \hfill
    \parbox{.8\linewidth}{\centering\small c. Estimation of $\xi_3$}
	\hfill
	\caption{Average RABias values (plus and minus one standard deviation) for 20 repeated experiments, as a function of the sample numbers $n$.}
\label{newfig1}
\end{figure}
\end{center}

\vspace{-3em}
The results are shown graphically in Figure~\ref{newfig}.  As expected, a degradation of the estimation performance is observed when the error probability $\rho$ increases using noisy and uncertain labels. But our solution based on soft labels does not suffer as much that using noisy labels, and it clearly outperforms the supervised learning with noisy labels. The estimations for $\xi_1$ and $\xi_3$ by our approach (uncertain labels) are better than the unsupervised learning with unknown labels. Although the estimation result for $\xi_2$ using uncertain labels seems not better than that by traditional EM algorithm when $\rho$ is large, it still indicates that our approach is able to exploit additional information on data uncertainty when such information is available as the case when $\rho$ is small.

In the following experiment, we will test the algorithm  with different sample numbers $n$. In order to illustrate the different behavior of the approach with respect to $n$, we consider a fixed censored scheme with $(m=)~60\%$ of samples are censored. With a given $n$, the test scheme is as follows: $m=n*0.6$,\linebreak $R=(0,0,\cdots,n-m)_{1 \times m}$. Let the error probability be $\rho=0.1$. Also we will compare our method using uncertain labels with those by noisy labels and without using any information of labels. The RABias for the results with different methods is shown in Figure~\ref{newfig1}. We can get similar conclusions as before that uncertainty on class labels appears to be successfully exploited by the proposed approach.  Moreover, as $n$ increases, the RABias  decreases, which indicates the large sample properties of the maximum-likelihood estimation.
\section{Conclusion}
In this paper, we investigate how to apply E2M algorithm to progressively censored data analysis. From the numerical results we can see that the proposed method based on E2M algorithm has a better behavior in terms of the RABias of the parameter estimations as it could take advantage of the available data uncertainty. Thus the belief function theory is an effective tool to represent and deal with the uncertain information in reliability evaluation. The Monte-Carlo simulations show that the RABiases decreases with the increase of $n$ for all cases. The method does improve for large sample size.

The mixture distribution is widely used in reliability project. Engineers find that there are often failures of tubes or other devices at the early stage, but the failure rate will remain stable or continue to raise
with the increase of time. From the view of statistics, these products should be regarded to come from mixed distributions. Besides, when the reliability evaluation of these complex products is performed, there is often not enough priori information. Therefore, the application of the proposed method is of practical meaning in this case.

\bibliographystyle{splncs03}
\bibliography{paperlist}
\end{document}